\documentclass[runningheads]{llncs}
\usepackage{marvosym}
\usepackage{graphicx}
\usepackage{amsmath}
\usepackage{amsfonts}
\usepackage{multirow}
\usepackage{color}

\usepackage{hyperref}
\hypersetup{
    colorlinks=true,
    linkcolor=blue,      
    urlcolor=blue,
    citecolor=blue,
}

\newcommand{\xmli}[1]{{\color[rgb]{0.9,0.1,0.3}{[XM:#1]}}}

\begin{document}
\title{FSDiffReg: Feature-wise and Score-wise Diffusion-guided Unsupervised Deformable Image Registration for Cardiac Images}
% \title{Efficient Unsupervised Deformable Medical Image Registration via Diffusion Model}
\author{Yi Qin and Xiaomeng Li$^($\textsuperscript{\Letter}$^)$}
% 1{Qin, Yi}
% 2{Li, Xiaomeng}{corresponding}
\institute{Department of Electronic and Computer Engineering, The Hong Kong University of Science and Technology, Hong Kong, China
\\ \email{eexmli@ust.hk}
}
\maketitle              % typeset the header of the contribution
\begin{abstract}
Unsupervised deformable image registration is one of the challenging tasks in medical imaging. Obtaining a high-quality deformation field while preserving deformation topology remains demanding amid a series of deep-learning-based solutions. Meanwhile, the diffusion model's latent feature space shows potential in modeling the deformation semantics. To fully exploit the diffusion model's ability to guide the registration task, we present two modules: Feature-wise Diffusion-Guided Module (FDG) and Score-wise Diffusion-Guided Module (SDG). Specifically, FDG uses the diffusion model's multi-scale semantic features to guide the generation of the deformation field. SDG uses the diffusion score to guide the optimization process for preserving deformation topology with barely any additional computation. Experiment results on the 3D medical cardiac image registration task validate our model's ability to provide refined deformation fields with preserved topology effectively. Code is available at: \url{https://github.com/xmed-lab/FSDiffReg.git}.
\end{abstract}

\section{Introduction}
Deformable image registration is the process of accurately estimating non-rigid voxel correspondences, such as the deformation field, between the same anatomical structure of a moving and fixed image pair. Fast, accurate, and realistic image registration algorithms are essential to improving the efficiency and accuracy of clinical practices. 
By observing dynamic changes, such as lesions, physicians can more comprehensively design treatment plans for patients~\cite{giger2013breast,jain2021amalgamation}.
When images during surgery align with preoperative ones, surgeons can locate instruments better and improve surgical prognosis~\cite{alam2018medical}. 
As reported in~\cite{khalil2018overview}, cardiac image registration is especially vital in improving heart chamber analysis accuracy, correcting cardiac imaging errors, and guiding cardiac surgeries. 
Thus, several studies have explored classical~\cite{klein2009elastix,avants2008symmetric} and deep-learning-based~\cite{balakrishnan2019voxelmorph,mok2020large,kim2021cyclemorph,huang2021difficulty} registration methods over the years.

% Accordingly, various works have studied the classical~\cite{klein2009elastix,avants2008symmetric} and deep-learning-based~\cite{balakrishnan2019voxelmorph,mok2020large,kim2021cyclemorph,huang2021difficulty} registration methods over the years.

Classical registration methods~\cite{klein2009elastix} used hand-crafted features to align images by solving computational-expensive optimization problems. Recently, researchers explored the deep-learning-based unsupervised deformable image registration~\cite{balakrishnan2019voxelmorph,kim2021cyclemorph,mok2020fast,mok2020large} to address the computational burden while reducing the need for accurate ground truth in the registration task. 
VoxelMorph~\cite{balakrishnan2019voxelmorph}, as the baseline, took moving and fixed image pairs as the input and maximized image pair similarity to train a registration network. To achieve higher accuracy, most unsupervised methods adopted a cascaded network with several sub-networks or an iterative refinement strategy~\cite{mok2020large,che2023amnet,huang2021difficulty,kim2021cyclemorph}. 
These strategies made the training procedure complicated and computational resources demanding. 
Meanwhile, to obtain smoother and more realistic deformation fields, i.e., topology preservation, many existing works introduced explicit diffeomorphic constraints~\cite{dalca2019unsupervised,mok2020fast,krebs2018unsupervised} or additional calculations on cycle consistency~\cite{kim2021cyclemorph}. For example, CycleMorph~\cite{kim2021cyclemorph} utilized the bidirectional registration consistency to preserve the topology during training. VoxelMorph-Diff~\cite{dalca2019unsupervised} adopted velocity field-based deformation field and new diffeomorphic estimation. SYMNet~\cite{mok2020fast} used symmetric deformation field estimation to achieve the goal. However, these schemes did not fully exploit the inherent network features, thereby overlooking these features' ability for better topology preservation.

%These schemes also need redundant calculation in the training process, therefore increasing the complexity and the need for computation resources.
%For instance, LapIRN[] utilized Laplacian feature pyramids to estimate the deformation field in multi resolution. AMNet[] estimated the local importance map and energy map in multiple scales to guide the generation of the deformation field. [2021MIA,diffculty aware] used multiple sub-networks to iterative refine the difficult patches of the registration input. 
\begin{figure}[b]
    \centering
    \includegraphics[width=0.85\textwidth]{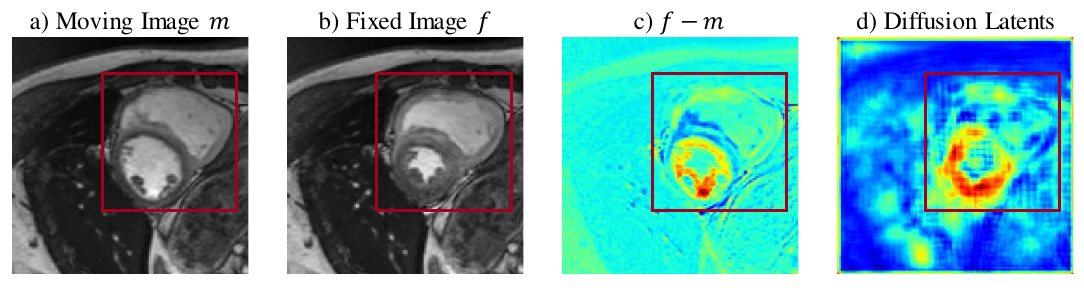}
    \caption{$f-m$ intuitively indicates where significant deformation occurs, as the red-boxed area shows. We calculated the per voxel energy score in the diffusion model's latent obtained in~\cite{kim2022diffusemorph} as~\cite{huang2021difficulty} suggests to identify the areas where complex deformation is likely to happen. The result indicates the same area, which was not explicitly utilized in prior work~\cite{kim2022diffusemorph}.}
    \label{fig:intro_fig}
\end{figure}

Recently, Kim~\textit{et al.}~\cite{kim2022diffusemorph} first proposed a diffusion model~\cite{ho2020denoising}, which is simpler to train than other generative models yet rich in semantics, for the registration task. They used the latent feature from the diffusion model's score function, i.e., the gradient field of a distribution's log-likelihood function~\cite{song2020score}, as one of the registration network's inputs for a better registration result. However, this method only used the final diffusion score as an image level guidance, which \textit{ignored diffusion model's rich task-specific semantics in the feature levels}, as proven in~\cite{kwon2023diffusion,baranchuk2022labelefficient,tumanyan2022plug}. 
%Therefore, the features learned at the hidden layers of the registration network could not be directly guided by the diffusion model's latent semantics, which reduced the informativeness of features for image registration. 
This resulted in the latent semantics of the diffusion model not being able to directly guide the features learned at the hidden layers of the registration network. As a result, the informativeness of these features for image registration was reduced. 
Moreover, this method only preserved deformation topology by simply using the diffusion score as the input, thereby \textit{ignoring the informative details about areas where significant deformations occur}; see Fig.~\ref{fig:intro_fig}.d for unexploited informative semantics. 
Therefore, the registration network was unable to explicitly prioritize hard-to-register areas, thereby limiting its effectiveness in preserving the deformation topology.
% Therefore, the registration network  could not explicitly focus on hard-to-register areas, limiting its ability to preserve the deformation topology effectively.

%We regard DiffuseMorph \cite{kim2022diffusemorph} as the baseline as it is the first and only work to integrate the score-based generative model with the registration task.

To address these issues, we present two novel modules, namely \textbf{F}eature-wise \textbf{D}iffusion-\textbf{G}uided Module (\textbf{FDG}) and \textbf{S}core-wise \textbf{D}iffusion-\textbf{G}uided Module (\textbf{SDG}) in the registration network. 
FDG introduces a direct feature-wise diffusion guidance technique for generating deformation fields by utilizing cross-attention to integrate the intermediate features of the diffusion model into the hidden layer of the registration network's decoder.
Furthermore, we embed the feature-wise guidance into multiple layers of the registration network and produce the feature-level deformation fields in multiple scales. 
Finally, after obtaining deformation fields at multiple scales, we upsample and average them to generate the full-resolution deformation field for registration.
%we combine the multiple feature-level deformation field to generate the final deformation field. \xmli{This sentence is confusing. }
Our SDG introduces explicit score-wise diffusion guidance for deformation topology preservation by reweighing the similarity-based unsupervised registration loss based on the diffusion score. Through this reweighing scheme, direct attention is given during the optimization process to ensure the preservation of the deformation topology. Our main contribution can be summarized as follows:

\begin{itemize}
\item We propose a novel feature-wise diffusion-guided module (FDG), which utilizes multi-scale intermediate features from the diffusion model to effectively guide the registration network in generating deformation fields. 
   
\item We also propose a score-wise diffusion-guided module (SDG), which leverages the diffusion model's score function to guide deformation topology preservation during the optimization process without incurring any additional computational burden. 

\item Experimental results on the cardiac dataset validated the effectiveness of our proposed method.
\end{itemize}

% more efficient -> converge faster, more stable, more compact, more accurate, multi-task way of doing this task.

%\input{Section2_RelatedWork}

\section{Method}
%\xmli{Given an input fixed image, we feed this image into xxxx, then we can a Warped image. Finally, by optimizing xxx loss function, we can obtain the final xxx. -- please follow the template to describe the whole process of Fig 2.}  

\subsection{Baseline Registration Model}

\begin{figure}[t]
    \centering
    \includegraphics[width=\textwidth]{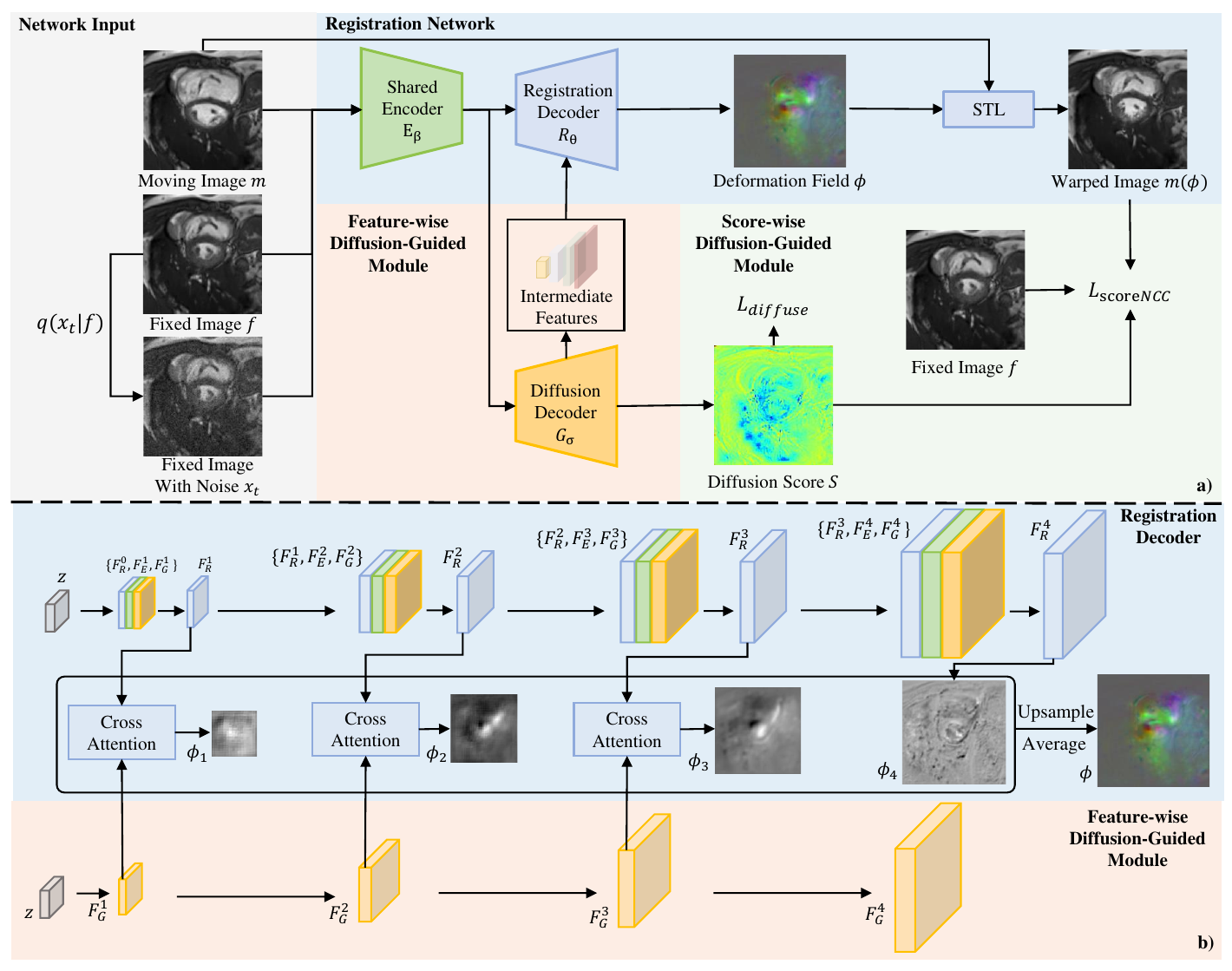}
    \caption{a) The workflow of FSDiffReg. b) The illustration of the guidance process of the Feature-wise Diffusion-Guided Module.}
    \label{fig:main_fig}
\end{figure}

Fig.~\ref{fig:main_fig}.a shows the overview of our proposed method. We first sample a perturbed noisy image $x_{t}$ from the fixed target image $f$ following the same scheme in~\cite{ho2020denoising}, which can be formulated as Equ.~\ref{eq:eq1}:
\begin{center}
\begin{equation}
\begin{aligned}
& x_{t}=\sqrt{\alpha_{t}}f+\sqrt{1-\alpha_t}\epsilon, \\ &\text{where } \alpha_t = {\prod^{t}_{s=1}}(1-\beta_s) \text{, } \epsilon \sim \mathcal{N}(0,\mathit{I})
\label{eq:eq1}
\end{aligned}
\end{equation}
\end{center}
where $0<\beta_s<1$ is the variance of the noise, $t$ is the noise level. 
Then we perform the registration training task. Given an input $x_{in}$ consisting of a fixed reference image $f$, a moving unaligned image $m$, and the perturbed noisy image $x_t$, we feed this input $x_{in}=\{f,m,x_t\}$ into the registration network's shared encoder $E_\beta$, followed by the registration decoder $R_\theta$. Then, the registration decoder $R_\theta$ outputs a deformation field $\phi$, guided by our \textit{Feature-wise Diffusion-Guided} module $G_\sigma$. Afterward, we feed $m$ and $\phi$ into the spatial transformation layer (STL) to generate the warped image $m(\phi)$. Finally, by optimizing the similarity-based loss function $L_{scoreNCC}$ guided by our \textit{Score-wise Diffusion-Guided} module, we can obtain the final registration model.

\subsection{Feature-wise Diffusion-Guided Module} 

The main component of the Feature-wise Diffusion-Guided module (FDG) is an auxiliary denoising diffusion decoder $G_\sigma$. The workflow of FDG is shown in Fig.~\ref{fig:main_fig}.b. Given the input $x_{in}=\{f,m,x_{t}\}$, the UNet shared encoder $E_\beta$ extracts the representation $z$. $z$ is then fed into the diffusion decoder $G$ and the registration decoder $R$ to get intermediate feature map pairs $F_i=\{(F_G^i,F_R^i)\},i=1,...,N$ from the $i$-th layer of the decoder. Of note, we generate the registration decoder's feature map by incorporating the guidance from the diffusion decoder, which can be formulated as Equ.~\ref{eq:layer}:

\begin{equation}
F_R^i=r_i(\mathrm{concat}(F_R^{i-1},F_E^{i},F_G^{i}))\text{, where }i=1,...,N
\label{eq:layer}
\end{equation}

where $r_i$ is the $i$-th layer of the registration decoder, and $F_E^{i}$ is the skip connection of features from the shared encoder layer at the same depth.

After obtaining the feature map pairs, our FDG module estimates the $i$-th feature level deformation field $\phi_i$ from the feature map pair $(F_G^i,F_R^i)$ using linear cross attention~\cite{shen2021efficient}, which can be defined as Equ.~\ref{eq:crossattn}:

\begin{equation}
\phi_i=\mathrm{Conv}(\mathrm{softmax}(F_R^i(\mathrm{GroupNorm}({F_G^i})^T \cdot F_G^i))+F_R^i)
\label{eq:crossattn}
\end{equation}

After obtaining all feature-level deformation fields from the shallowest layer to the deepest layer, we generate the final deformation field $\phi$ by enlarging and averaging all feature-level deformation fields. This is a commonly adopted method for multi-scale deformation field merging so as to merge features which attend different scales and granularity. The final $\phi$ is then fed into the spatial transformation layer with the moving image $m$ to generate the registered image $m(\phi)$.

\subsection{Score-wise Diffusion-Guided Module}

Given the representation $z$ encoded by the shared encoder $z=E_\beta(x_{in})$, the diffusion decoder $G_\sigma$ outputs a diffusion score estimation $S=G_\sigma(z)$. Then, the Score-wise Diffusion-Guided Module (SDG) uses this score to reweigh the similarity-based normalized cross-correlation loss function, formulated as Equ.~\ref{eq:scorenccloss}:

\begin{equation}
L_{scoreNCC}(m,f,S)=(\frac{1}{1+e^{-S}})^\gamma \odot -(m(\phi) \otimes f)
\label{eq:scorenccloss}
\end{equation}

where $m(\phi)$ is the warped moving image, $\odot$ defines the Hadamard product, and $\otimes$ defines the local normalized cross-correlation function. $\gamma$ is a hyperparameter to amplify the reweighing effect.

By this means, SDG utilizes the diffusion score to explicitly indicate the hard-to-register areas, i.e., areas where deformation topology is hard to preserve, then assigning higher weights in the loss function for greater attention, and vice versa for easier-to-register areas. Therefore, the information on deformation topology is effectively incorporated into the optimization process without additional constraints by the SDG module.

\subsection{Overall Training and Inference}

\subsubsection{Loss function.}
Our network predicts the deformable fields at the feature level and then outputs the registered image. The total loss function of our method is defined as Equ.~\ref{eq:totalloss}:

\begin{equation}
 L_{total}=L_{diffusion}(x_{in},t)+\lambda L_{scoreNCC}(m,f,S)+\lambda_\phi\sum{||\nabla_\phi||^2}
\label{eq:totalloss}
\end{equation}

\begin{equation}
L_{diffusion}(x_{in},t)=\mathbb{E}_z{||G_\sigma(E_\beta(x_{in},t))-\epsilon||^2_2}\text{ ,where }\epsilon \sim \mathcal{N}(0,\mathit{I})
\label{eq:diffusionloss}
\end{equation}

where $L_{diffusion}$ is the auxiliary loss function for training the diffusion decoder $G_\sigma$ (Equ.~\ref{eq:diffusionloss}), and $t$ is the noise level of $x_t$, following the method in~\cite{ho2020denoising}. Our proposed $L_{scoreNCC}$ encourages maximizing the similarity between the registered and reference images while preserving the deformation topology. $\sum{||\nabla_\phi||^2}$ is the conventional smoothness penalty on the deformation field. $\lambda$ and $\lambda_\phi$ are hyperparameters, and we empirically set them to $20$ in our experiments.

\subsubsection{Inference.}

In the inference stage, we perform image registration in the same style as~\cite{kim2022diffusemorph}. Instead of the perturbed image $x_t$, we input the original reference image $f$ into the network, and the total network input becomes $x_{in}=\{f,m,f\}$. Given this network input $x_{in}$, our network first generates the deformation field $\phi$ between the moving image $m$ and the reference image $f$ and produces the registered moving image $m(\phi)$ by feeding the moving image $m$ and the deformation field $\phi$ into the spatial transformation layer (STL). The registered moving image is the final output of our network.

\section{Experiments and Results}

\subsubsection{Dataset and Preprocessing}

Following the previous work~\cite{kim2022diffusemorph}, we used the publicly available 3D cardiac MR dataset ACDC~\cite{bernard2018deep} for experiments. The dataset includes 100 4D temporal cardiac MRI data with corresponding segmentation maps. We selected the 3D image at the end of the diastolic stage as the fixed image and the image at the end of the systolic stage as the moving image. 
We resampled all scans to the voxel spacing of $1.5\times1.5\times3.15 mm$, then cropped them to the voxel size of $128\times128\times32$. We normalized the intensity of all images to $\lbrack-1,1\rbrack$. The training set contains 90 image pairs, while the remaining 10 pairs form the test set.
The abovementioned preprocessing steps were performed in accordance with the approach described in prior work~\cite{kim2022diffusemorph} to ensure a fair comparison.

\subsubsection{Implementation Details}

The proposed framework was implemented using the PyTorch library, version 1.12.0.
Following~\cite{kim2022diffusemorph}, we used DDPM UNet's 3D encoder as our shared encoder and DDPM UNet's 3D decoder as our diffusion decoder. For the registration part, instead of a complete 3D UNet in~\cite{kim2022diffusemorph}, we only used DDPM UNet's 3D decoder as our registration decoder to generate the deformation field.
During the diffusion task, we gradually increased the noise schedule from $10^{-6}$ to $10^{-2}$ over 2000 timesteps.
We utilized an Nvidia RTX3090 GPU and the Adam optimization algorithm~\cite{kingma2014adam} to train the model with $\lambda=20$, $\lambda_\phi=20$, $\gamma=1$, batch size $\mathbb{B}=1$, a learning rate of $2\times10^{-4}$, and a maximum of 700 epochs.

\subsubsection{Evaluation Metrics}
We employed three evaluation metrics, i.e., DICE, $|J|\leq0(\%)$, and $SD(|J|)$ to measure the image registration performance, following existing registration methods~\cite{balakrishnan2019voxelmorph,kim2022diffusemorph,kim2021cyclemorph}. 
DICE measures the spatial overlap of anatomical segmentation maps between the warped moving image and the fixed reference image.
A higher Dice score indicates better alignment between the warped moving image and the fixed reference image, thus reflecting an improved registration quality.
$|J|\leq0(\%)$ indicates the percentage of non-positive values in the Jacobian determinant of the registration field. 
This metric indicates the percentage of voxels that lacks a one-to-one registration mapping relation,  causing unrealistic deformations and roughness. 
$SD(|J|)$ refers to the standard deviation of the Jacobian determinant of the registration field. A lower standard deviation indicates that the registration field is relatively smooth and consistent across the image.

\subsubsection{Compare with the State-of-the-art Methods}
\begin{figure}[htb]
    \centering
    \includegraphics[width=\textwidth]{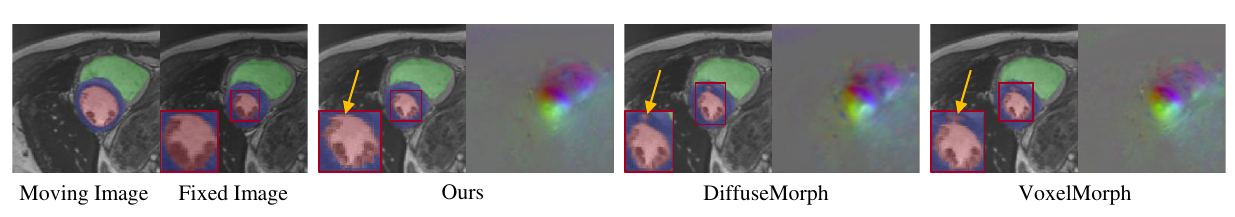}
    \caption{The visualization of registration results. As the red-boxed area shows, our deformation field and the corresponding image are more refined in the area where larger deformation happens.}
    \label{fig:compare_fig}
\end{figure}

Table~\ref{tab1} shows the comparison of our method with existing state-of-the-art methods including VoxelMorph~\cite{balakrishnan2019voxelmorph}, VoxelMorph-Diff~\cite{dalca2019unsupervised}, and DiffuseMorph~\cite{kim2022diffusemorph} on the same training and testing dataset. We produced baseline results using the recommended hyperparameters in their paper.
%we present the average Dice score for the segmentation masks of the myocardium(Myo), left ventricle(LV), right ventricle(RV), and the overall result. 
The result shows that our proposed method outperforms existing baseline methods by a substantial margin (Wilcoxon signed-rank test, $p<0.005$) (Also see Fig.~\ref{fig:compare_fig}). Furthermore, our method aligned better in areas where larger deformation happened, such as myocardium (myo).
%The average runtime of our method is 0.581s, which is on par with other deep-learning-based algorithms~\cite{balakrishnan2019voxelmorph,kim2022diffusemorph}.
%It demonstrates that guided by the diffusion model's feature-level semantics, the registration decoder can obtain more informative features for generating deformation fields, thus getting better registration accuracy (Also see Fig.~\ref{fig:attn_fig}). \xmli{this conclusion can't be obtained by seeing Table 1.} The results also show that guided by diffusion scores, the network's optimization process can better concentrate on hard-to-register areas, thus better preserving the deformation topology. \xmli{The same}

\begin{table}[!h]
\centering
\caption{Image registration results with standard deviation in parenthesis on the 3D cardiac dataset. ``LV'', ``Myo'', ``RV'' refers to Left Ventricle, Myocardium, and Right Ventricle, respectively. ``Overall'' refers to the averaged registration result of the left blood pool, myocardium, left ventricle, right ventricle, and these total region, following~\cite{kim2022diffusemorph}. $\uparrow$: the higher, the better results. $\downarrow$: the lower, the better results.}\label{tab1}
\resizebox{\textwidth}{!}{%
\begin{tabular}{c|cccc|cc}
\hline
\multirow{2}{*}{Method}                                 & \multicolumn{4}{c|}{DICE $\uparrow$}                                               & \multirow{2}{*}{$|J|\leq0(\%) \downarrow$} & \multirow{2}{*}{$SD(|J|) \downarrow$}\\ \cline{2-5}
                                       & LV           & Myo                & RV           & Overall                                           \\ \hline
Initial                                & 0.585(0.074) & 0.357(0.120)  & 0.741(0.069) & 0.655(0.188)          & - & -                       \\
VM~\cite{balakrishnan2019voxelmorph}    & 0.770(0.086) & 0.679(0.129)  & 0.816(0.065) & 0.799(0.110)          & 0.079(0.058)&0.183             \\
VM-Diff~\cite{dalca2019unsupervised}    & 0.755(0.092) & 0.659(0.137)  & 0.815(0.066) & 0.789(0.117)          & 0.083(0.063)&0.182             \\
DiffuseMorph~\cite{kim2022diffusemorph} & 0.783(0.086) & 0.678(0.148)  & 0.821(0.067) & 0.805(0.114)          & 0.061(0.038)&0.178             \\
\textbf{Ours}                          & \textbf{0.809(0.077)} & \textbf{0.724(0.119)}  & \textbf{0.827(0.061)} & \textbf{0.823(0.096)} & \textbf{0.054(0.026)}&\textbf{0.176}    \\ \hline
\end{tabular}%
}
\end{table}

\begin{figure}[htb]
    \centering
    \includegraphics[width=\textwidth]{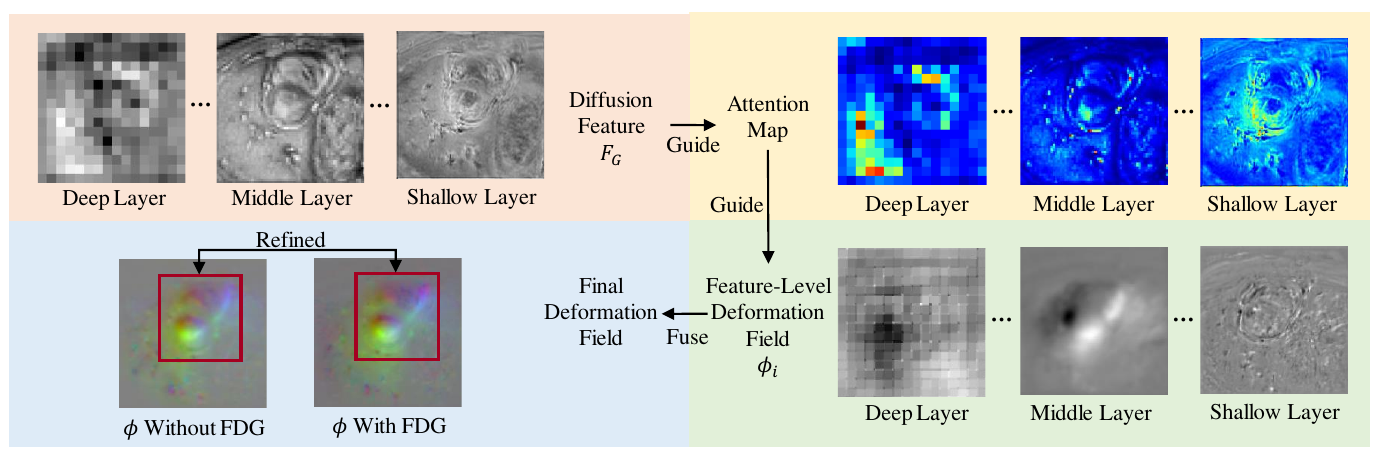}
    \caption{The visualization example of the effectiveness of FDG. The deformation field shows more details and is more distinct when guided by the FDG, thus improving registration accuracy, as the area in the red box shows.}
    \label{fig:attn_fig}
\end{figure}

\subsubsection{Ablation Study}
To validate the effectiveness of our proposed learning strategies, including the Feature-wise Diffusion-Guided module(FDG) and Score-wise Diffusion-Guided module(SDG), we conducted ablative experiments, as shown in Table~\ref{tab2}. The network without FDG also uses the denoising diffusion decoder but generates the deformation field from the encoded feature directly, and without SDG means that we optimize the network using the vanilla NCC loss. By integrating multi-scale intermediate latent diffusion features into generating deformation fields, we can see that the network's performance increased by 1\%. By deploying the reweighing loss, the Jacobian metric decreased by 60.5\%. The result achieved a balance when all components were deployed. These results demonstrated that our proposed components could effectively guide the deformation field generation by using multi-scale diffusion features (Also see Fig.~\ref{fig:attn_fig}). Optimization guided by diffusion score led to better preservation of deformation topology. It is worth noticing that the results without FDG or SDG showed only marginal improvement over baseline results, indicating the importance of feature-level deformation field generation and the reweighing scheme. The ablative study of hyperparameter $\lambda$ is illustrated in Supp. Fig. 1.

\begin{minipage}{0.95\textwidth}
\begin{minipage}[t]{0.48\textwidth}
\makeatletter\def\@captype{table}
%\begin{table}[htb]
\centering
\caption{Ablation study on FDG and SDG.}\label{tab2}
\resizebox{0.9\textwidth}{!}{%
\begin{tabular}{cc|cc}
\hline
FDG & SDG & DICE$\uparrow$ & $|J|\leq0(\%)\downarrow$ \\ \hline
                                   &                 &0.811(0.098)      &0.114(0.076)   \\
              $\surd$                    &                 & 0.818(0.102)     & 0.062(0.038)  \\
                                   & $\surd$               & 0.810(0.188)     & 0.045(0.025)  \\
 $\surd$                    & $\surd$               & 0.823(0.096) & 0.054(0.026)   \\\hline
\end{tabular}%
}
%\end{table}
\end{minipage}
\begin{minipage}[t]{0.48\textwidth}
\makeatletter\def\@captype{table}
\centering
\caption{Ablation Study on hyperparameter $\gamma$ in SDG.}\label{tab3}
\resizebox{\textwidth}{!}{%
\begin{tabular}{c|ccc}
\hline
$\gamma$ & 0.5 & 1 & 2 \\ \hline
DICE$\uparrow$                  &0.817(0.100)     &0.823(0.096)   &0.816(0.104)  \\
$|J|\leq0(\%)\downarrow$                     &0.069(0.029)     &0.054(0.026)   &0.042(0.023)  \\ \hline
\end{tabular}%
}
\end{minipage}
\end{minipage}

%lambda, exponential.

\subsubsection{Analysis of $\gamma$}
%\xmli{This para shows the new table with analysis of $\lambda$.} 
Furthermore, to validate SDG's effectiveness on topology preservation, we conducted another ablative study on SDG's hyperparameter $\gamma$, as Table~\ref{tab3} shows. Increased $\gamma$ indicates a more substantial reweighing effect. The results showed that by adding stronger reweighing influence, we could obtain deformation fields with better topology preservation almost without compromising accuracy.
\iffalse
\begin{table}[htb]
\centering
\caption{Ablation Study on hyperparameter $\gamma$}\label{tab3}
\resizebox{0.48\textwidth}{!}{%
\begin{tabular}{c|ccc}
\hline
$\gamma$ & 0.5 & 1 & 2 \\ \hline
DICE$\uparrow$                  &0.817(0.100)     &0.823(0.096)   &0.816(0.104)  \\
$|J|\leq0(\%)\downarrow$                     &0.069(0.029)     &0.054(0.026)   &0.042(0.023)  \\ \hline
\end{tabular}%
}
\end{table}
\fi

\iffalse
\begin{table}[]
\label{tab:tab1}
\begin{tabular}{cccccc}
\hline
Method        & DICE$\uparrow$                  & $|J|\leq0(\%)\downarrow$                   & PSNR(dB)$\uparrow$               & NMSE$\downarrow$                  & Time$\downarrow$                  \\ \hline
Initial       & 0.655(0.188)          & -                     & 23.813(2.852)          & 0.223                 & -                     \\
VM            & 0.781(0.110)          & 0.167(0.109)          & 28.092(4.850)          & 0.105(0.084)          & 0.561(1.259)          \\
VM-Diff       & 0.793(0.103)          & 0.284(0.188)          & 27.977(4.694)          & 0.163(0.072)          & 0.636(1.264)          \\
DiffuseMorph  & 0.809(0.117)          & 0.107(0.091)          & 28.105(3.545)          & 0.091(0.061)          & \textbf{0.438(1.126)} \\
\textbf{Ours} & \textbf{0.815(0.100)} & \textbf{0.066(0.037)} & \textbf{29.493(4.026)} & \textbf{0.072(0.055)} & 0.535(1.309) \\ \hline         
\end{tabular}
\end{table}
\fi

\section{Conclusion}
This work proposes two novel modules for unsupervised deformable image registration: the Feature-wise Diffusion-Guided module (FDG) and the Score-wise Diffusion-Guided module (SDG). Among these modules, FDG can effectively guide the deformation field generation by utilizing the multi-scale intermediate diffusion features. SDG demonstrates its ability to guide the optimization process for better deformation topology preservation using the diffusion score. Extensive experiments show that the proposed framework brings impressive improvements over all baselines. The proposed work models the non-linear deformation semantics using the diffusion model. Therefore, it is sound to generalize to other registration tasks and images, which may be one of the future research directions.

\noindent \textbf{Acknowledgements.}
This work was supported by the Hong Kong Innovation and Technology Fund under Project ITS/030/21 \& PRP/041/22FX, as well as by Foshan HKUST Projects under Grants FSUST21-HKUST10E and FSUST21-HKUST11E.

\bibliographystyle{splncs04}
\bibliography{paper1165}

\begin{thebibliography}{10}
\providecommand{\url}[1]{\texttt{#1}}
\providecommand{\urlprefix}{URL }
\providecommand{\doi}[1]{https://doi.org/#1}

\bibitem{alam2018medical}
Alam, F., Rahman, S.U., Ullah, S., Gulati, K.: Medical image registration in
  image guided surgery: Issues, challenges and research opportunities.
  Biocybernetics and Biomedical Engineering  \textbf{38}(1),  71--89 (2018)

\bibitem{avants2008symmetric}
Avants, B.B., Epstein, C.L., Grossman, M., Gee, J.C.: Symmetric diffeomorphic
  image registration with cross-correlation: evaluating automated labeling of
  elderly and neurodegenerative brain. Medical image analysis  \textbf{12}(1),
  26--41 (2008)

\bibitem{balakrishnan2019voxelmorph}
Balakrishnan, G., Zhao, A., Sabuncu, M.R., Guttag, J., Dalca, A.V.: Voxelmorph:
  a learning framework for deformable medical image registration. IEEE
  transactions on medical imaging  \textbf{38}(8),  1788--1800 (2019)

\bibitem{baranchuk2022labelefficient}
Baranchuk, D., Voynov, A., Rubachev, I., Khrulkov, V., Babenko, A.:
  Label-efficient semantic segmentation with diffusion models. In:
  International Conference on Learning Representations (2022),
  \url{https://openreview.net/forum?id=SlxSY2UZQT}

\bibitem{bernard2018deep}
Bernard, O., Lalande, A., Zotti, C., Cervenansky, F., Yang, X., Heng, P.A.,
  Cetin, I., Lekadir, K., Camara, O., Ballester, M.A.G., et~al.: Deep learning
  techniques for automatic mri cardiac multi-structures segmentation and
  diagnosis: is the problem solved? IEEE transactions on medical imaging
  \textbf{37}(11),  2514--2525 (2018)

\bibitem{che2023amnet}
Che, T., Wang, X., Zhao, K., Zhao, Y., Zeng, D., Li, Q., Zheng, Y., Yang, N.,
  Wang, J., Li, S.: Amnet: Adaptive multi-level network for deformable
  registration of 3d brain mr images. Medical Image Analysis p. 102740 (2023)

\bibitem{dalca2019unsupervised}
Dalca, A.V., Balakrishnan, G., Guttag, J., Sabuncu, M.R.: Unsupervised learning
  of probabilistic diffeomorphic registration for images and surfaces. Medical
  image analysis  \textbf{57},  226--236 (2019)

\bibitem{giger2013breast}
Giger, M.L., Karssemeijer, N., Schnabel, J.A.: Breast image analysis for risk
  assessment, detection, diagnosis, and treatment of cancer. Annual review of
  biomedical engineering  \textbf{15},  327--357 (2013)

\bibitem{ho2020denoising}
Ho, J., Jain, A., Abbeel, P.: Denoising diffusion probabilistic models.
  Advances in Neural Information Processing Systems  \textbf{33},  6840--6851
  (2020)

\bibitem{huang2021difficulty}
Huang, Y., Ahmad, S., Fan, J., Shen, D., Yap, P.T.: Difficulty-aware
  hierarchical convolutional neural networks for deformable registration of
  brain mr images. Medical image analysis  \textbf{67},  101817 (2021)

\bibitem{jain2021amalgamation}
Jain, M., Rai, C., Jain, J., Gambhir, D.: Amalgamation of machine learning and
  slice-by-slice registration of mri for early prognosis of cognitive decline.
  International Journal of Advanced Computer Science and Applications
  \textbf{12}(1) (2021)

\bibitem{khalil2018overview}
Khalil, A., Ng, S.C., Liew, Y.M., Lai, K.W.: An overview on image registration
  techniques for cardiac diagnosis and treatment. Cardiology research and
  practice  \textbf{2018} (2018)

\bibitem{kim2022diffusemorph}
Kim, B., Han, I., Ye, J.C.: Diffusemorph: Unsupervised deformable image
  registration using diffusion model. In: Computer Vision--ECCV 2022: 17th
  European Conference, Tel Aviv, Israel, October 23--27, 2022, Proceedings,
  Part XXXI. pp. 347--364. Springer (2022)

\bibitem{kim2021cyclemorph}
Kim, B., Kim, D.H., Park, S.H., Kim, J., Lee, J.G., Ye, J.C.: Cyclemorph: cycle
  consistent unsupervised deformable image registration. Medical image analysis
   \textbf{71},  102036 (2021)

\bibitem{kingma2014adam}
Kingma, D.P., Ba, J.: Adam: A method for stochastic optimization. arXiv
  preprint arXiv:1412.6980  (2014)

\bibitem{klein2009elastix}
Klein, S., Staring, M., Murphy, K., Viergever, M.A., Pluim, J.P.: Elastix: a
  toolbox for intensity-based medical image registration. IEEE transactions on
  medical imaging  \textbf{29}(1),  196--205 (2009)

\bibitem{krebs2018unsupervised}
Krebs, J., Mansi, T., Mailh{\'e}, B., Ayache, N., Delingette, H.: Unsupervised
  probabilistic deformation modeling for robust diffeomorphic registration. In:
  Deep Learning in Medical Image Analysis and Multimodal Learning for Clinical
  Decision Support: 4th International Workshop, DLMIA 2018, and 8th
  International Workshop, ML-CDS 2018, Held in Conjunction with MICCAI 2018,
  Granada, Spain, September 20, 2018, Proceedings 4. pp. 101--109. Springer
  (2018)

\bibitem{kwon2023diffusion}
Kwon, M., Jeong, J., Uh, Y.: Diffusion models already have a semantic latent
  space. In: The Eleventh International Conference on Learning Representations
  (2023), \url{https://openreview.net/forum?id=pd1P2eUBVfq}

\bibitem{mok2020fast}
Mok, T.C., Chung, A.: Fast symmetric diffeomorphic image registration with
  convolutional neural networks. In: Proceedings of the IEEE/CVF conference on
  computer vision and pattern recognition. pp. 4644--4653 (2020)

\bibitem{mok2020large}
Mok, T.C., Chung, A.C.: Large deformation diffeomorphic image registration with
  laplacian pyramid networks. In: Medical Image Computing and Computer Assisted
  Intervention--MICCAI 2020: 23rd International Conference, Lima, Peru, October
  4--8, 2020, Proceedings, Part III 23. pp. 211--221. Springer (2020)

\bibitem{shen2021efficient}
Shen, Z., Zhang, M., Zhao, H., Yi, S., Li, H.: Efficient attention: Attention
  with linear complexities. In: Proceedings of the IEEE/CVF winter conference
  on applications of computer vision. pp. 3531--3539 (2021)

\bibitem{song2020score}
Song, Y., Sohl-Dickstein, J., Kingma, D.P., Kumar, A., Ermon, S., Poole, B.:
  Score-based generative modeling through stochastic differential equations.
  arXiv preprint arXiv:2011.13456  (2020)

\bibitem{tumanyan2022plug}
Tumanyan, N., Geyer, M., Bagon, S., Dekel, T.: Plug-and-play diffusion features
  for text-driven image-to-image translation. arXiv preprint arXiv:2211.12572
  (2022)

\end{thebibliography}
\end{document}